\documentclass[11pt]{article}
\usepackage{float}

\usepackage[final]{acl}

\usepackage{times}
\usepackage{latexsym}

\usepackage[T1]{fontenc}

\usepackage[utf8]{inputenc}
\usepackage{makecell}
\usepackage{microtype}

\usepackage{inconsolata}

\usepackage{graphicx}

\usepackage{pdfpages}  
\usepackage{longtable} 
\usepackage{tabularx}
\usepackage{booktabs}
\usepackage{array}
\usepackage{graphicx} 
\usepackage{makecell}  
\usepackage{siunitx}
\usepackage{multirow,arydshln}
\usepackage{todonotes}

\everypar{\looseness=-1}

\title{Linear Semantic Segmentation for Low-Resource Spoken Dialects}

\author{
Kirill Chirkunov\textsuperscript{1},
Younes Samih\textsuperscript{2},
Abed Alhakim Freihat\textsuperscript{1},
Hanan Aldarmaki\textsuperscript{1} \\
\textsuperscript{1}Mohamed bin Zayed University of Artificial Intelligence \\
\textsuperscript{2}IBM Research AI \\
\texttt{\{kirill.chirkunov, abdelhakim.freihat, hanan.aldarmaki\}@mbzuai.ac.ae} \\
\texttt{younes.samih@ibm.com}
}

\begin{document}
\maketitle
\begin{abstract}


Semantic segmentation is a core component of discourse analysis, yet existing models are primarily developed and evaluated on high-resource written text, limiting their effectiveness on low-resource spoken varieties. In particular, dialectal Arabic exhibits informal syntax, code-switching, and weakly marked discourse structure that challenge standard segmentation approaches. In this paper, we introduce a new multi-genre benchmark (more than 1000 samples) for semantic segmentation in conversational Arabic, focusing on dialectal discourse. The benchmark covers transcribed casual telephone conversations, code-switched podcasts, broadcast news, and expressive dialogue from novels, and was annotated and validated by native Arabic annotators. Using this benchmark, we show that segmentation models performing well on MSA news genres degrade on dialectal transcribed speech. We further propose a segmentation model that targets local semantic coherence and robustness to discourse discontinuities, consistently outperforming strong baselines on dialectal non-news genres. The benchmark and approach generalize to other low-resource spoken languages.

\end{abstract}

\section{Introduction}
Modern natural language processing systems are typically developed under an implicit structural assumption that text is organized into well-defined and coherent units. In tasks such as long-context summarization and Retrieval-Augmented Generation (RAG), models rely on structural cues -- including paragraph boundaries, standardized punctuation, and syntactic regularity -- to infer semantic structure and determine discourse boundaries \citep{hearst-1997-text, lukasik-etal-2020-text}. When such cues are present, as in Wikipedia articles or news-wire text, current models perform well. However, recent work shows that when these cues are absent, segmentation quality degrades substantially, leading to cascading failures in downstream retrieval and generation tasks \citep{Ghinassi2024}.

This breakdown is particularly pronounced in \textbf{Dialectal Arabic}. Unlike Modern Standard Arabic (MSA), which adheres to relatively consistent orthographic and grammatical conventions, dialectal Arabic often appears in transcripts of conversational speech and thus exhibits many conversational-language traits: non-standard and inconsistent spelling \citep{Zaidan2014}, frequent code-switching \citep{Hamed2024}, dense colloquial morphology, and weakly marked discourse boundaries. As a result, informal Arabic speech -- such as podcasts, telephone dialogues, or talk shows -- poses a different segmentation problem from the formal text on which most NLP systems are trained. In these settings, conventional indicators of topic transition are often obscured by non-standard variations and the absence of reliable punctuation \citep{Ghosh2022}.

While substantial progress has been made on sentence- and token-level tasks, including dialect identification, sentiment analysis, and named entity recognition \citep{Bouamor2018, Darwish2021}, discourse-level modeling for informal Arabic remains underdeveloped. Existing resources can often identify dialect IDs, but not when a speaker transitions between topics or narrative segments. This limitation restricts the applicability of higher-level NLP systems, such as semantic search, content structuring, and long-form analysis, for the majority of transcribed spoken Arabic content. 
Most dialectal Arabic resources, including MADAR, Shami, and Curras, are organized as isolated utterances, reflecting their design for sentence-level classification tasks \citep{Bouamor2018, AbuKwaik2018, Jarrar2016Curras}. At the same time, large collections of spoken and code-switched Arabic speech remain unsegmented, forcing downstream systems to rely on heuristic chunking strategies that fragment semantic coherence \citep{ali-mgb-5-2019, al-ali-aldarmaki-2024-mixat}. Although recent Arabic-focused large language models and multilingual embeddings improve generative fluency and lexical coverage \citep{jais2-2025, bari2025allam}, they do not directly address discourse segmentation under structural discontinuity and informal speech conditions \citep{bhatia-etal-2025-swan, dunn-2023-variation}.

\emph{Linear semantic segmentation} (also called \emph{linear text segmentation}) is the task of splitting a text into contiguous segments, where each segment is semantically coherent and addresses a specific topic. For example, a phone call may begin with the initial topic and then shift to other topics. A linear semantic segmentation model is trained to detect suitable boundaries between segments:
\begin{quote}
\small
\noindent\textbf{Example (phone call).}\\[1mm]

\begin{tabular}{@{}r p{0.82\linewidth}@{}}
1 & Salam, I am calling about my Grand Prix tickets. \\
2 & I got the confirmation email, but the QR code is not opening with me. \\
3 & No problem, can you give me the booking number? \\
4 & Yes, okay, I found your order in the system. \\
5 & I will resend the tickets now, you will get them in a minute. \\
6 & Perfect, thanks. Also, for Gate 6, where should I park? \\
7 & For Gate 6, best to use West Parking. It is the closest one. \\
\end{tabular}\\[1mm]

\textit{Segmentation:}\\
segment\#1: lines 1,2,3,4,5, topic: \emph{ticket issue} \\
segment\#2: lines 6,7, topic: \emph{parking information}
\end{quote}

In this work, we introducing a multi-genre benchmark specifically designed to evaluate semantic segmentation in low-resource, spoken Arabic. Prior segmentation models largely assume gradual topic transitions and global discourse coherence, assumptions that break down in spoken and expressive dialectal Arabic. Recent work on dialect-sensitive embeddings shows that Arabic dialects induce distinct semantic subspaces \citep{dunn-2023-variation}, further complicating discourse modeling. Building on this insight, we propose a new segmentation model that explicitly targets local semantic coherence and robustness to dialectal variation and discourse discontinuities. We evaluate our model against other standard and state-of-the-art semantic segmentation models across multiple non-news genres and show consistent improvements over strong baselines in dialectal and informal settings. 
Our contribution in this work can be summarized as follows:
\begin{enumerate}
    \item We release the first open-source dataset providing gold-standard semantic segmentation for Dialectal Arabic, covering diverse and underrepresented genres, including casual telephone conversations, code-switched podcasts, and expressive literary dialogue.
    \item We conduct a systematic evaluation of classical, neural, and large language model–based segmentation approaches, demonstrating that methods that achieve strong performance on MSA news degrade sharply on dialectal inputs, regardless of model scale.
    \item We propose a domain-adaptive segmentation model that prioritizes local semantic coherence over global structural cues, yielding consistent improvements for the noisy and spontaneous discourse patterns characteristic of spoken Arabic.
\end{enumerate}

The data and model are available online\footnote{\url{https://github.com/mbzuai-nlp/DialSeg-Ar}}.

\raggedbottom
\section{Related Work}
Semantic segmentation, defined as the identification of boundaries between topically coherent discourse units, is a long-standing problem in natural language processing and discourse analysis \citep{hearst-1997-text, purver2011topic}. Early approaches relied on lexical cohesion and distributional shifts, while recent neural models have achieved strong performance on structured, written English by leveraging hierarchical encoders and Transformer-based architectures \citep{lukasik-etal-2020-text, Glavas2021}. These models, however, implicitly depend on editorial regularities such as standardized punctuation, paragraph boundaries, and consistent sentence structure. 

A growing body of work demonstrates that segmentation performance degrades substantially when models trained on written text are applied to spoken or quasi-spoken language, including meetings, call-center interactions, and casual chat \citep{Ghosh2022, Zhong2022}. Spoken discourse is structurally unstable, characterized by disfluencies, interruptions, speaker overlap, and implicit topic shifts. Recent benchmarks such as YTSEG \citep{Retkowski2024} represent important progress toward evaluating segmentation on spoken English. 
Despite substantial progress in Arabic NLP, semantic segmentation for Arabic remains largely unexplored beyond Modern Standard Arabic (MSA). Existing Arabic segmentation resources are either confined to formal MSA domains, such as broadcast news within the Topic Detection and Tracking framework \citep{LDC2006T20}, or focus on sentence-level and morphological annotation in dialectal corpora. Large-scale dialectal resources such as MADAR, Shami, Curras, and related lexicons \citep{Bouamor2018, AbuKwaik2018, Jarrar2016Curras} have been instrumental for tasks such as dialect identification and sentiment analysis, but consist primarily of isolated utterances and lack discourse-level structural annotation. 
A substantial volume of dialectal and code-switched Arabic speech data exists without topic-level segmentation, including multi-genre broadcast speech such as MGB-5 \citep{ali-mgb-5-2019}, conversational telephone speech from the CallHome and CallFriend corpora \citep{ldc2006t15,ldc2006t16,ldc2007t01}, and more recent Arabic--English code-switched datasets such as MixAt and ZAEBUC \citep{al-ali-aldarmaki-2024-mixat, Hamed2024}. 
Our benchmark complements these resources by providing golden semantic segmentation that enables controlled evaluation of segmentation models under realistic dialectal and code-switched conditions.

Recent advances in large language models and embedding-based representations have renewed interest in semantic segmentation. Arabic-focused models such as Jais and ALLAM \citep{jais2-2025, bari2025allam}, alongside multilingual embedding models such as Gemma \citep{embedding-gemma-2025, gemma_2025}, offer strong general-purpose representations. However, recent evidence suggests that embedding-based segmentation methods remain sensitive to noise, syntactic variability, and orthographic inconsistency, particularly in dialectal settings \citep{bhatia-etal-2025-swan, alwajih-etal-2025-palm}. Our systematic evaluation confirms these findings, showing that models performing well on news deteriorate sharply on dialectal speech, even when powered by modern embedding and LLM-based infrastructures.



\section{Dataset}

\subsection{Source corpora and genres}

Table~\ref{tab:domains_corpora} summarizes all domains, source corpora, genres, and references used in our benchmark. Below, we only highlight properties that are most relevant for semantic segmentation.


{\textbf{MGB-5:}} For Moroccan Arabic, we use the publicly available MGB-5 multi-genre broadcast corpus \citep{ali-mgb-5-2019,arabicspeech-mgb5-hf-2025}. Transcripts contain orthographic variation, giving us realistic, noisy dialectal text. 


{\textbf{LDC:}} To broaden the coverage of spoken dialectal Arabic, we incorporate three LDC conversational telephone corpora in Gulf, Iraqi, and Levantine Arabic \citep{ldc2006t15,ldc2006t16,ldc2007t01}. In line with LDC licensing,\footnote{\url{https://www.ldc.upenn.edu/data-management/using/licensing}} our released benchmark only includes references to the original files blocks, not the underlying texts.


{\textbf{Podcasts:}} For code-switching, we rely on the {Mixat} Gulf Arabic-English podcast corpus \citep{al-ali-aldarmaki-2024-mixat}. We use randomly picked consistent blocks from each podcast episode as a source for segmentation.


{\textbf{Rewayat:}} To approximate spoken Gulf Arabic dialogue, we extract samples of short conversations from dialectical novels published on the Web\footnote{\url{https://www.rewity.com/}}. We lightly normalize them and annotate speaker turns.



{\textbf{OPUS:}} As a standard written baseline, we include Modern Standard Arabic (MSA) news commentary texts from OPUS \citep{tiedemann-2012-parallel}. This well-structured source serves as a baseline to measure performance against other parts of the benchmark.

\begin{table*}[ht]
\centering
\small
\renewcommand{\arraystretch}{1.3}
\scalebox{0.9}{
\begin{tabular}{p{0.08\textwidth}p{0.20\textwidth}p{0.30\textwidth}p{0.15\textwidth}p{0.15\textwidth}}
\hline
\textbf{Domain} & \textbf{Source corpora / Dialect} & \textbf{Description} & \textbf{Genre(s)} & \textbf{References} \\ \hline
\multirow{2}{*}{Clean text} 
  & {\textbf{OPUS}} News Commentaries (Modern Standard Arabic) 
  & Well structured, clean news texts from OPUS (parallel corpus). A traditional benchmark for text segmentation. 
  & economy, sports, politics 
  & \citep{tiedemann-2012-parallel} \\ \cline{2-5}
  & {\textbf{Rewayat}} conversations (Gulf) 
  & Extracted conversations from dialectical novels. Written expressive/figurative speech. 
  & drama, history 
  & Web sources \\ \hline
\multirow{2}{*}{Noisy text} 
  & {\textbf{MGB-5}}, Multi-genre broadcast transcripts (Moroccan Arabic) 
  & Manually annotated YouTube programs. Short phrases with allowed orthographic variation. 
  & comedy, cooking, family shows, fashion, drama, science, sports 
  & \citep{ali-mgb-5-2019,arabicspeech-mgb5-hf-2025} \\ \cline{2-5}
  & {\textbf{LDC}} conversational telephone speech transcripts (Gulf/Iraqi/Levantine) 
  & Spontaneous two-party colloquial phone conversations transcripts. 
  & casual 
  & \citep{ldc2006t15,ldc2006t16,ldc2007t01} \\ \hline
{Code switching} 
  & {\textbf{Podcasts}} transcripts, Gulf-English speech from Mixat
  & Podcasts recorded in high quality and transcribed (2 shows × 14 episodes). 
  & finance, health, technology 
  & \citep{al-ali-aldarmaki-2024-mixat,sqrk-mixat-tri-hf-2025} \\ \hline
\end{tabular}
}
\caption{Overview of source corpora by domain, dialect, description, genres, and references.}
\label{tab:domains_corpora}
\end{table*}

\subsection{Data preparation}

As a first step, we convert all corpora into a unified conversational representation. Each text is mapped to a linear list of utterances with an associated speaker ID. For naturally conversational data (telephone speech, podcasts, written dialogue), we preserve existing turn boundaries. For news, we use sentence segmentation and treat each sentence as one utterance. This representation allows us to apply a single segmentation protocol across spoken, quasi-spoken, and written genres.

We then obtain initial synthetic segmentation using the gpt-oss-120b model. For each text (utterance sequence), we prompt the model to build topic-boundary blocks\footnote{The prompt used is provided in Appendix \ref{sec:appendix-syn-annotation-prompt}} following the  task representation used for semantic segmentation in \citet{fan-etal-2024-uncovering}. These machine-generated boundaries serve purely as proposals. After this step, texts are split into train, validation, and test sets, stratified by language variety and genre. Only the validation and test partitions receive full human verification and correction as described in the next section. The training split may use synthetic boundaries and/or partially verified data depending on the experiment.

\flushbottom
\subsection{Human Annotation}
Human annotation is organized into two subtasks: within-segment and cross-segment validation. In the within-segment validation subtask, annotators inspect each proposed segment and flag utterances that are clearly off-topic. 
In the cross-segment validation subtask, they review the list of consecutive segments and merge adjacent segments that express the same topic.  
Together, the two validation tasks correct both boundary insertion and boundary deletion errors. The annotation guidelines are provided in Appendix \ref{sec:appendix_guidelines}.

For each dialect / genre, two independent native Arabic annotators produce initial corrections, and a third annotator acts as the adjudicator, resolving disagreements and ensuring consistency. 
Before large-scale annotation, adjudicators participated in a pilot phase to test and refine the guidelines. The final guidelines were then used in a short training session for annotators, focusing on workflow, edge cases, and expected output. After annotation, adjudicators manually inspected disagreements, making targeted corrections where necessary. The released benchmark thus contains a clean set of gold segmentation, with human-verified splits for evaluation and carefully controlled synthetic labels for training.

A comparison silver-vs-golden annotated data is provided in Appendix~\ref{sec:appendix-silver-vs-gold-comparison}. Only 21\% of silver annotations were changed during gold annotation, meaning that about four out of five silver annotations were retained, with source-level change rates ranging from 14\% (\textit{Rewayat}) to 31\% (\textit{OPUS}).

\subsection{Quality Check}
Inter-annotator agreement scores for all datasets, including both topic- and line-level Po, Cohen’s $\kappa$, and Gwet’s AC1, are summarized in Table~\ref{tab:iaa}.

\begin{table}[ht]
\centering
\small
\renewcommand{\arraystretch}{1.1}
\scalebox{0.9}{
\begin{tabular}{lcccccc}
\hline
\textbf{Dataset} 
& \multicolumn{3}{c}{\textbf{Cross-segment}} 
& \multicolumn{3}{c}{\textbf{Within-segment}} \\ 
& Po & $\kappa$ & AC1 & Po & $\kappa$ & AC1 \\ \hline
OPUS (MSA) & 0.87 & 0.74 & 0.83 & 1.00 & 0.00 & 1.00 \\
Rewayat (GL) & 0.84 & 0.71 & 0.76 & 0.94 & 0.02 & 0.93 \\
MGB-5 (MR) & 0.88 & 0.72 & 0.84 & 1.00 & 0.00 & 1.00 \\
LDC (GL) & 0.83 & 0.67 & 0.70 & 0.83 & 0.01 & 0.77 \\
LDC (LV) & 0.79 & 0.65 & 0.67 & 0.91 & 0.37 & 0.90 \\
LDC (IQ) & 0.81 & 0.71 & 0.73 & 0.85   & 0.71  & 0.99 \\
Podcasts (CS) & 0.81 & 0.62 & 0.72 & 0.98 & 0.49 & 0.98 \\ \hline
\end{tabular}
}
\caption{Inter-annotator agreement for cross-segment and within-segment annotation across datasets: observed agreement (Po), Cohen's $\kappa$, and Gwet's AC1.}
\label{tab:iaa}
\end{table}

\subsection{Descriptive Statistics}
Descriptive statistics are presented in Table~\ref{tab:dataset_statistics}. 

\begin{table*}[ht]
\centering
\setlength{\tabcolsep}{3pt}
\renewcommand{\arraystretch}{1.1}
\scalebox{0.8}{
\begin{tabular}{lccccccccccccc}
\hline
\textbf{Dataset} & \textbf{Toks.} 
& \multicolumn{3}{c}{\textbf{Toks. / Utt.}} 
& \textbf{Utts.} 
& \multicolumn{3}{c}{\textbf{Utts. / Sample}} 
& \textbf{Segs.} 
& \multicolumn{3}{c}{\textbf{Segs. / Sample}} 
& \textbf{Samples}
\\ \cline{3-5} \cline{7-9} \cline{11-13}
 &  & Avg & Min & Max &  & Avg & Min & Max &  & Avg & Min & Max &  \\ \hline
OPUS (MSA) & 153,234 & 22.21 & 1 & 99 & 6,899 & 28.16 & 8 & 54  & 681 & 2.78 & 1 & 9 & 245 \\
Rewayat (GL) & 58,222 & 9.29 & 1 & 84 & 6,269 & 25.08 & 21 & 55  & 988 & 3.95 & 1 & 14 & 250 \\
MGB-5 (MR) & 11,609 & 9.15 & 1 & 34 & 1,269 & 57.68 & 50 & 116 & 123 & 5.59 & 2 & 13 & 22 \\
LDC (GL) & 63,872 & 6.57 & 1 & 36 & 9,718 & 56.50 & 31 & 107 & 770 & 4.48 & 1 & 17 & 172 \\
LDC (LV) & 49,165 & 4.64 & 1 & 36 & 10,591 & 67.03 & 32 & 217 & 605 & 3.83 & 1 & 14 & 158 \\
LDC (IQ) & 46,726 & 4.99 & 1 & 31 & 9,355 & 72.52 & 27 & 156 & 570 & 4.12 & 1 & 18 & 129 \\
Podcasts (CS) & 35,252 & 19.85 & 1 & 49 & 1,776 & 52.24 & 33 & 61 & 163 & 4.79 & 1 & 11 & 34 \\ \hline
\textbf{Overall} & \textbf{371,354} & \textbf{10.17} & \textbf{1} & \textbf{99} & \textbf{36,522} & \textbf{41.46} & \textbf{8} & \textbf{217} & \textbf{3,900} & \textbf{3.86} & \textbf{1} & \textbf{18} & \textbf{1,010} \\ \hline
\end{tabular}}
\caption{Descriptive statistics across datasets. Columns show the dataset subset, total tokens (Toks.), tokens per utterance (Toks./Utt.), total utterances (Utts.), utterances per sample (Utts./Sample), total segments (Segs.), segments per sample (Segs./Sample), and the number of samples; grouped columns report average, minimum, and maximum values.}
\label{tab:dataset_statistics}
\end{table*}

\section{Dialectical Semantic Segmentation model}
\subsection{Design rationale}
Recent work suggests that large language models are promising tools for discourse-level tasks such as semantic segmentation and dialogue structuring, but their potential is only beginning to be explored systematically. \citet{fan-etal-2024-uncovering} shows that gpt-3.5-turbo can recover reasonably plausible topic structures in dialogue, yet also highlight clear shortcomings in domain-specific conversations. In parallel, survey work on linear text segmentation emphasizes that most existing systems still rely on comparatively shallow architectures and traditional resources, with limited coverage of LLM-based approaches and virtually no focus on low-resource, multilingual discourse settings \citep{ghinassi-etal-2024-recent}. This motivates exploration of modern LLMs as a backbone for semantic segmentation in low-resource language settings rather than relying solely on classical or sentence-level segmenters.

For dialectal Arabic, there is growing evidence that multilingual and Arabic-centric multilingual models are preferable to purely monolingual Arabic encoders. Cross-lingual transfer studies demonstrate that Arabic NLU benefits from shared representations with English and French, both via multilingual pretraining and translation-based augmentation \citep{abboud-etal-2022-cross}. Work on language adherence in multilingual LLMs further shows that it is possible to improve output-language control while retaining cross-lingual task transfer \citep{rahmati-etal-2025-coco}. Recent Arabic-centric LLMs such as ALLaM and Jais-2 explicitly embrace bilingual or trilingual setups (Arabic–English, or Arabic–English–French+code), and report consistent gains across downstream tasks compared to monolingual baselines \citep{bari2025allam,jais2-2025}. At the same time, new community models like NileChat and Atlas-Chat, as well as Darija-focused adaptations such as GemMaroc, illustrate that dialectal capabilities emerge most strongly when multilingual foundations are combined with targeted dialect-specific data and objectives \citep{el-mekki-etal-2025-nilechat,shang-etal-2025-nile,shang-etal-2025-atlas,skiredj2025gemmaroc,altakrori2025dialectalarabicmmlu}. These trends suggest that a multilingual, instruction-tuned model with a multilingual background is a better starting point for dialectal segmentation than traditional monolingual models (e.g., AraBERT-style encoders) \citep{antoun2020arabert}.

Within this framework, we choose Gemma-3-4b-it \citep{gemma_2025} as our base model. Gemma family models offer competitive multilingual performance and are increasingly adopted in dialect-focused systems and benchmarks, including recent work on Moroccan and Egyptian Arabic LLMs \citep{el-mekki-etal-2025-nilechat, shang-etal-2025-nile, skiredj2025gemmaroc, shang-etal-2025-atlas, altakrori2025dialectalarabicmmlu}. Crucially for discourse analysis, Gemma-3 supports context windows up to 128K tokens, allowing us to process long conversations, podcast episodes, and multi-page documents in a single pass -- something encoder-only models like SaT \citep{frohmann-etal-2024-segment} cannot do due to their much shorter maximum context length (typically 512 tokens) and more limited subword vocabularies.

\subsection{Training process}

We adopt a supervised setup in which synthetic semantic segmentations produced by gpt-oss-120b (Section~3.2) are used as supervision labels for fine-tuning a model, while human-corrected segmentations are reserved for validation and test sets. 
To avoid overfitting to high-resource subsets (e.g., OPUS news commentaries or LDC corpus), we use stratified sampling over both genre and dialect when forming the training set. 


To make the model domain-adaptive and oriented toward local semantic coherence rather than global topic cues, we explicitly condition the training prompts on both genre and dialect via the \texttt{data\_source} and \texttt{language\_clue} fields in the prompt and define boundaries solely in terms of topic shifts between neighboring utterances (see Appendix~\ref{sec:appendix-segmentation-prompt}, Appendix~\ref{sec:appendix-restoration-prompt}). This setup encourages the learned representations to generalize across heterogeneous formats while remaining sensitive to local discourse changes in noisy, spontaneous dialectal speech.

To improve robustness, we add an auxiliary segmentation restoration task on top of standard segmentation. 
Following the UL2 approach \citep{ul2}, we treat the primary segmentation task as an \emph{extreme} denoising setting and the restoration task as an \emph{easier} denoising setting.
We create corrupted variants by randomly merging adjacent segments. The primary segmentation task predicts boundaries from scratch given a sequence of utterances. The restoration task takes a corrupted segmentation of the same utterance sequence and recovers the boundaries. Both tasks share the same representation, and we train with a joint loss that sums cross-entropy over clean and corrupted examples, encouraging the model to both segment and repair noisy discourse structures.
Corruption is applied with a skewed distribution over merged spans (1:60\%, 2:20\%, 3:15\%, 4:5\%). It encourages the model to segment from scratch or repair noisy discourse structures, improving generalization and robustness.


\section{Experimental Setup}
We  finetuned Gemma3-4B with LoRA adapters using the Hugging Face ecosystem: transformers, peft and trl libraries. All experiments were conducted on a single node equipped with 4×A100 GPUs (80GB each), with training running for roughly 500 steps and completing in about two hours per model.

\subsection{Modeling params}
In LLMs for structured JSON output, we use near-deterministic settings (temperature = 0 / 0.1) to improve format compliance and reduce run-to-run variance. We checked that performance is insensitive to small temperature changes in tests.
In other methods, we use the recommended default parameter values.

\subsection{Evaluation}

We evaluate all models on the test split, stratified by genre and dialect, as described in Section~3. Predictions are compared to gold segmentations at the utterance level: for each utterance boundary position, the model produces a binary boundary/non-boundary decision. We report\textbf{ macro F1} for boundary detection and topic accuracy, defined as the proportion of utterances whose segment membership matches the gold segmentation after aligning predicted and gold segments.
However, macro F1 is not the best primary metric for linear text segmentation because segmentation errors are structured and distance-sensitive, while F1 is not \citep{pevzner-hearst-2002-critique}.
For more robust evaluation, we compute\textbf{ Pk} and \textbf{WindowDiff (WD)}, which are more tolerant than an exact match and are able to capture structural effects based on a sliding window.

To study the effect of language and script on segmentation quality, we perform a cross-lingual analysis on a representative subset of documents. Using Google Translate, we derive (i) MSA versions of dialectal texts and (ii) English translations of Arabic texts while keeping document structure intact. We then run baselines and our model on three parallel versions of each document (original dialectal Arabic, MSA, and English) and compare utterance-level F1, Pk, and WindowDiff (WD) across languages. This setup isolates the impact of language and resource availability on segmentation performance while holding underlying discourse content fixed.

\subsection{Baselines}

\subsubsection{Unsupervised baselines}

As classical unsupervised baselines, we use TextTiling \citep{hearst-1997-text} and C99 \citep{choi-2000-advances}, alongside their Arabic adaptations, ArabTextTiling and ArabC99, proposed for Arabic news semantic segmentation \citep{chaibi2014,Naili2018TheCO}. All methods rely on lexical cohesion scores over adjacent text blocks but differ in how similarity matrices are normalized and smoothed. 

To incorporate stronger semantic similarity signals, we implement a TextTiling-style segmenter where lexical similarity is replaced by embedding-based similarity between adjacent utterance blocks, following the TeT+CLS idea of \citet{Xu_Zhao_Zhang_2021}. We consider two embedding models: (i) TeT+CLS\textsubscript{DA}: a MARBERTv2-based SentenceTransformer trained for Arabic and dialectal semantic similarity \citep{abdul-mageed-etal-2021-arbert, nacar2024enhancingsemanticsimilarityunderstanding}, and (ii) TeT+CLS\textsubscript{Multi}: EmbeddingGemma, which provides lightweight multilingual representations  \citep{embedding-gemma-2025}. 

\subsubsection{Supervised baseline}

As a strong supervised baseline, we use the  Segment Any Text (SaT) multilingual model \citep{frohmann-etal-2024-segment}, built on XLM-R and trained to produce robust sentence/paragraph boundaries across noisy domains. We map our utterances to SaT’s input units (sentences or short segments), run the model on each text to get paragraphs split, and project its predicted boundaries back to the utterance level. 


\subsubsection{LLM-based baselines}

Finally, we evaluate instruction-following Arabic-centric LLMs as text segmenters. We consider NileChat-12B, a model with explicit support for low-resource Arabic dialects \citep{el-mekki-etal-2025-nilechat}, ALLaM-7B-Instruct, a bilingual Arabic–English model designed to handle Arabic spoken dialects \citep{bari2025allam}, Fanar-1-9B-Instruct \citep{fanarllm2025}, and multilingual Gemma3-4B. All models are prompted with the same discourse-oriented segmentation instructions as our model (Section~4) and are prompted to output splits over utterances. We apply them directly to the test set without task-specific fine-tuning and evaluate using the same metrics (macro F1, Pk, WD). 

\section{Results}

\begin{table*}[t]
\centering
\footnotesize
\setlength{\tabcolsep}{2.6pt}
\renewcommand{\arraystretch}{1.15}

\newcommand{\vtype}[1]{\rotatebox[origin=c]{90}{\textnormal{ #1}}}
\newcommand{\best}[1]{\multicolumn{1}{S[table-format=1.2]}{\bfseries #1}}

\scalebox{0.8}{
\begin{tabular}{>{\centering\arraybackslash}p{2.2mm}  
                >{\raggedright\arraybackslash}p{1.9cm} 
                *{15}{S[table-format=1.2]}}
\toprule
\textbf{} & \textbf{Method}
& \multicolumn{3}{c}{\textbf{OPUS (MSA)}}
& \multicolumn{3}{c}{\textbf{Rewayat (GL)}}
& \multicolumn{3}{c}{\textbf{MGB-5 (MR)}}
& \multicolumn{3}{c}{\textbf{\shortstack{LDC (GL/IQ/LV)}}}
& \multicolumn{3}{c}{\textbf{Podcasts (CS)}} \\
\cmidrule(lr){3-5}\cmidrule(lr){6-8}\cmidrule(lr){9-11}\cmidrule(lr){12-14}\cmidrule(lr){15-17}
& & {F1($\uparrow$)} & {Pk($\downarrow$)} & {WD($\downarrow$)} 
  & {F1($\uparrow$)} & {Pk($\downarrow$)} & {WD($\downarrow$)} 
  & {F1($\uparrow$)} & {Pk($\downarrow$)} & {WD($\downarrow$)} 
  & {F1($\uparrow$)} & {Pk($\downarrow$)} & {WD($\downarrow$)} 
  & {F1($\uparrow$)} & {Pk($\downarrow$)} & {WD($\downarrow$)}  \\
\midrule

\multirow{6}{*}{\vtype{Unsupervised}}
& TextTiling
& 0.47 & 0.68 & 0.75 & 0.48 & 0.52 & 0.53 & 0.50 & 0.55 & 0.71 & 0.47 & 0.57 & 0.68 & 0.48 & 0.63 & 0.78 \\
& AraTextTiling
& 0.47 & 0.50 & 0.54 & 0.50 & 0.47 & 0.49 & 0.50 & 0.52 & 0.57 & 0.49 & 0.49 & 0.54 & 0.50 & \best{0.49} & 0.56 \\
& C99
& 0.49 & 0.55 & 0.60 & 0.44 & 0.49 & 0.55 & 0.44 & 0.50 & 0.60 & 0.48 & 0.51 & 0.56 & 0.48 & 0.53 & 0.63 \\
& ArabC99
& 0.51 & 0.54 & 0.57 & 0.48 & 0.47 & 0.51 & 0.49 & 0.53 & 0.60 & 0.48 & 0.51 & 0.56 & 0.50 & 0.54 & 0.62 \\
& \makecell[l]{TeT+CLS\textsubscript{DA}}
& 0.58 & 0.44 & 0.47 & \best{0.56} & 0.43 & 0.44 & 0.51 & 0.55 & 0.60 & \best{0.52} & 0.50 & 0.57 & \best{0.54} & 0.51 & 0.57 \\
& \makecell[l]{TeT+CLS\textsubscript{Multi}}
& 0.50 & 0.48 & 0.51 & 0.49 & 0.46 & 0.47 & 0.48 & 0.51 & 0.57 & 0.49 & 0.52 & 0.58 & 0.52 & 0.50 & 0.56  \\
\midrule
& SaT {\scriptsize [supervised]}
& 0.60 & 0.50 & 0.62 & 0.28 & 0.61 & 0.97 & 0.49 & 0.50 & 0.63 & 0.38 & 0.58 & 0.82 & 0.50 & 0.56 & 0.72 \\
\midrule

\multirow{5}{*}{\vtype{LLMs}}
& \makecell[l]{ALLaM-7B-It}
& 0.56 & 0.57 & 0.65 & 0.53 & 0.54 & 0.62 & 0.45 & 0.55 & 0.67 & 0.48 & 0.60 & 0.77 & 0.51 & 0.61 & 0.81 \\
& \makecell[l]{Fanar-1-9B-It}
& 0.58 & 0.48 & 0.57 & 0.52 & 0.51 & 0.62 & 0.51 & 0.51 & 0.54 & 0.49 & 0.56 & 0.67 & 0.53 & 0.59 & 0.70  \\
& \makecell[l]{NileChat-12B}
& 0.57 & 0.53 & 0.57 & 0.55 & 0.46 & 0.48 & 0.51 & 0.53 & 0.57 & 0.50 & 0.60 & 0.69 & 0.53 & 0.56 & 0.64 \\
& \makecell[l]{Gemma3-4B}
& 0.49 & 0.66 & 0.75 & 0.54 & 0.55 & 0.59 & \best{0.54} & 0.52 & 0.65 & 0.50 & 0.49 & 0.52 & 0.49 & 0.56 & 0.67 \\
& \makecell[l]{[Ours]}
& \best{0.60} & \best{0.43} & \best{0.46} & 0.52 & \best{0.43} & \best{0.43} & 0.53 & \best{0.40} & \best{0.40} & 0.50 & \best{0.41} & \best{0.39} & 0.48 & 0.52 & \best{0.53} \\
\bottomrule
\end{tabular}}
\caption{Segmentation performance across five datasets. Higher is better for F1; lower is better for Pk and WD.}
\label{tab:segmentation_all}

\end{table*}

\begin{figure*}[t]
  \centering
  \includegraphics[width=\linewidth]{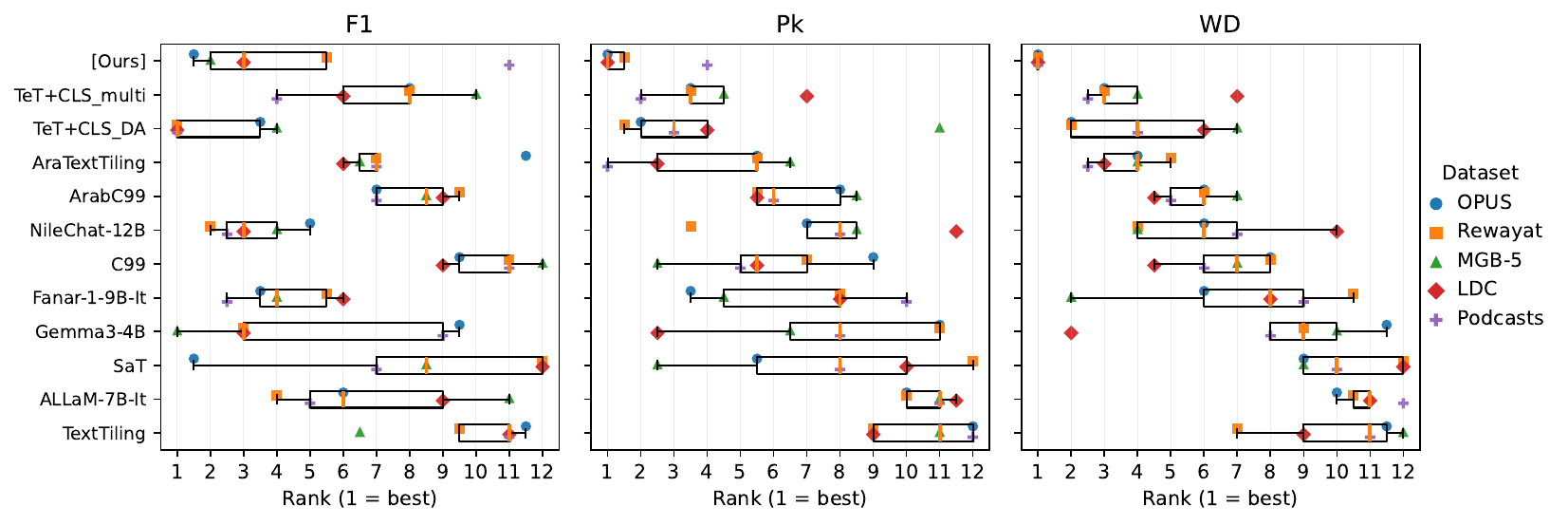}
  \caption{For each model, the box plot summarizes its rank distribution across five datasets, where the center line denotes the median rank, the box spans the interquartile range (IQR), i.e. 25th to 75th percentiles, and the whiskers extend from the box to the farthest data point lying within 1.5 x IQR from the box.}
  \label{fig:my_pdf}
\end{figure*}

The experimental results across all models and datasets are shown in Table ~\ref{tab:segmentation_all}. We also show the rank distribution of all models in Figure ~\ref{fig:my_pdf}. The figure illustrates how consistently each model ranks in comparison to all other models. We can see from the figure that, in terms of the fine-grained Pk and WD metrics, our approach leads to consistent improvements, as it ranks higher than other models. In terms of macro F1, the ranking distributions are less clear and more scattered, as the metric is coarser and insensitive to within-sequence performance variations. 

\paragraph{Performance on clean text (news, conversations from novels):} 

On clean, well-structured OPUS News (MSA) our model surprisingly shows the best performance over F1/Pk/WD. For Rewayat (GL), methods that explicitly model cohesion based on sentence embeddings shows good performance at the utterance level, as demonstrated in Macro F1 score; TeT+CLS\textsubscript{DA} achieved the best F1 score on this dataset, but in terms of Pk and WD, our model performs on a par with it.
Supervised and LLM baselines behave quite differently. The SaT model, trained on predominantly MSA, remains competitive on news (only in terms of F1) but degrades sharply on all dialectal corpora, showing that supervised encoders trained on standard high-resource languages do not necessarily transfer well to spoken dialects. 

\paragraph{Performance on  transcriptions of spoken dialectals (MGB-5, LDC):}

On spoken transcriptions, which tend to be noisy and  weakly structured, our approach outperforms all others in Pk and WD metrics, and is comparable to the highest performance in terms of F1. 
It builds segments more consistently despite highly irregular turn lengths, truncated sentences, and spontaneous topic shifts.
In contract, all other models show substantial degradations in terms of Pk and WD metrics compared to their performance on written text. 

\paragraph{Performance on  code-switching annotated podcast transcriptions:}

On the code-switched podcasts, segmentation becomes even harder: turns freely mix Gulf Arabic and English. Most models exhibit a higher error rate than on clean news or novels (cf. Table~\ref{tab:segmentation_all}). Our method achieves the lowest WD and competitive Pk, which indicates more consistent boundary placement, even when its F1 is lower than that achieved by TeT+CLS\textsubscript{DA} .

\paragraph {Performance after Machine Translation (Original -> English, English -> MSA):}

Across original and machine-translated settings (Table~\ref{tab:mt_comparison}), our model is the only model that consistently delivers low segmentation error in terms of Pk and WD across all settings. Other systems: unsupervised, embedding-based, and generic LLMs show modest fluctuations around their original Pk/WD values. The main exception is SaT, which is highly sensitive to translation. Overall, the results indicate that our method is generalizable across language variations.

\begin{table*}[ht]
\centering
\small
\renewcommand{\arraystretch}{1.3}
\scalebox{0.9}{
\begin{tabular}{llccccccccc}
\hline
\textbf{Type} & \textbf{Method}
& \multicolumn{3}{c}{\textbf{Original Language}}
& \multicolumn{3}{c}{\textbf{EN (MT)}}
& \multicolumn{3}{c}{\textbf{EN(MT)$\rightarrow$MSA(MT)}} \\ \cline{3-11}
 &  & F1($\uparrow$) & Pk($\downarrow$) & WD($\downarrow$)
    & F1($\uparrow$) & Pk($\downarrow$) & WD($\downarrow$)
    & F1($\uparrow$) & Pk($\downarrow$) & WD($\downarrow$) \\ \hline
\multirow{2}{*}{Unsupervised}
  & TextTiling        
  & 0.47 & 0.58 & 0.66 
  & 0.48 & 0.58 & 0.68 
  & \textbf{0.50} & \textbf{0.51} & \textbf{0.56} \\
  & AraTextTiling     
  & \textbf{0.49} & \textbf{0.49} & \textbf{0.53} 
  & -- & -- & -- 
  & 0.48 & 0.58 & 0.63 \\ 
  & C99     
  & \textbf{0.47} & \textbf{0.51} & \textbf{0.57} 
  & \textbf{0.47} & \textbf{0.51} & \textbf{0.57} 
  & 0.45 & 0.52 & 0.59 \\
  & ArabC99     
  & \textbf{0.49} & \textbf{0.51} & \textbf{0.55} 
  & -- & -- & -- 
  & \textbf{0.49} & \textbf{0.51} & \textbf{0.55} \\ 
  \hline
\multirow{2}{*}{Embeddings}
  & \makecell[l]{TeT+CLS$_{DA}$}     
  & 0.54 & \textbf{0.47} & \textbf{0.51} 
  & -- & -- & -- 
  & \textbf{\textit{0.55}} & \textbf{0.47} & 0.52 \\
  & \makecell[l]{TeT+CLS$_{Multi}$} 
  & \textbf{0.49} & 0.50 & 0.54 
  & \textbf{0.49} & \textbf{0.48} & \textbf{0.52} 
  & \textbf{0.49} & \textbf{0.48} & \textbf{0.52} \\ \hline
Supervised
  & SaT  
  & 0.41 & 0.56 & 0.80 
  & \textbf{0.49} & \textbf{0.52} & \textbf{0.63} 
  & 0.39 & 0.60 & 0.88 \\ \hline
\multirow{3}{*}{LLMs}
  & ALLaM-7B-It 
  & \textbf{0.52} & \textbf{0.57} & \textbf{0.68} 
  & 0.49 & 0.59 & 0.72 
  & 0.49 & 0.60 & 0.76 \\
  & Fanar-1-9B-It 
  & \textbf{0.52} & \textbf{0.52} & \textbf{0.62} 
  & 0.50 & 0.54 & 0.68 
  & 0.49 & 0.56 & 0.68 \\
  & Nile-Chat-12B  
  & \textbf{0.53} & \textbf{0.53} & \textbf{0.59} 
  & 0.52 & 0.54 & 0.63 
  & \textbf{0.53} & 0.54 & 0.66 \\
  & Gemma3-4B 
  & \textbf{0.51} & \textbf{0.55} & \textbf{0.60} 
  & 0.49 & 0.57 & 0.63 
  & 0.49 & 0.57 & 0.63 \\
  & [Ours] 
  & \textbf{0.52} & 0.43 & 0.43 
  & 0.51 & \textbf{\textit{0.42}} & \textbf{\textit{0.42}} 
  & \textbf{0.52} & 0.43 & 0.44 \\
  \hline
\end{tabular}
}
\caption{Segmentation performance (macro average) in the original language and under machine translation to high-resource language settings. Bold indicates the best performance within each method row, while bold italic indicates the best overall performance.}
\label{tab:mt_comparison}
\end{table*}

\subsection{Ablation studies}

\begin{table}[ht]
\centering
\small
\renewcommand{\arraystretch}{1.3}
\begin{tabular}{lccc}
\hline
\textbf{Ablation Experiments} & F1($\uparrow$) & Pk($\downarrow$) & WD($\downarrow$) \\ \hline
No finetuning & 0.50 & 0.49 & 0.55 \\
Finetuning w/o corruption & 0.49 & 0.46 & 0.47 \\
Finetuning w/corruption & {\textbf{0.52}} & {\textbf{0.43}} & {\textbf{0.43}}
\\ \hline
\end{tabular}
\caption{Ablation study on model finetuning methods. Macro average over all test subsets.}
\label{tab:ablation_prompt}
\end{table}

As shown in Table~\ref{tab:ablation_prompt}, the base Gemma3-4B without finetuning reaches F1=0.50 with Pk=0.49 and WD=0.55, computed as a macro average over all test subsets. Finetuning only on segmentation without the corruption–restoration task improves boundary placement (Pk=0.46/WD=0.47) but does not increase F1. Adding the corruption–restoration task yields the best trade-off (F1=0.52, Pk/WD=0.43), indicating that robustness to noisy segmentations directly translates into cleaner overall discourse partitions across our diverse data sources.

In addition, a complementary data ablation study in Appendix~\ref{sec:appendix-data-ablation-study} shows that the final model benefits from the diversity of the full multi-genre training mixture rather than from any single source alone.

\section{Discussion}

Our results indicate that open-source multilingual LLMs can provide a substantial boost for linear text segmentation over challenging cases in low-resource settings, but only once they are properly finetuned. A dialect-aware Gemma3-based segmenter not only outperforms traditional supervised models on dialectal and noisy data, but also remains robust across language channels (original, EN MT, MSA MT), unlike MSA-trained encoders, whose performance collapses off-domain. At the same time, the experiments also show the limits of relying on "out-of-the-box" LLMs: instruction prompting techniques and few-shot examples don't show significant improvement in comparison with traditional methods, and far from finetuned models. In practice, performance improvement in dialectal semantic segmentation comes rather from supervised adaptation for language variety.

\section {Conclusion}

We have shown that a carefully constructed benchmark and a targeted finetuning strategy are both necessary to make progress on semantic segmentation in dialectal, conversational settings. The benchmark itself, spanning multiple genres and dialects, exposes substantial performance gaps between clean MSA text and realistic transcriptions and reveals that MSA-trained models and off-the-shelf LLMs are not sufficient when code-switching, disfluencies, and language with orthographic variations are present. Within this framework, our finetuning methodology over Gemma3-4B segmenter provides a consistent gain in structural quality: it achieves the best Pk/WindowDiff scores on the most challenging dialectal and noisy subsets as well as on clean news and novels. Together, the benchmark and the finetuned model form a practical reference point for future work on discourse-level analysis in dialectical Arabic and, by extension, other low-resource spoken languages.

\section*{Limitations}

Our study is based exclusively on text transcripts and written texts. We do not use the underlying audio and therefore ignore prosody, pauses, turn overlap, and other acoustic cues that often signal topic shifts in spontaneous speech. It also does not take into consideration ASR errors that may appear in automatically transcribed transcripts. 
Moreover, our modeling objective is boundary-centric and evaluated using intrinsic segmentation metrics only. We are not measuring effect of segmentation for downstream tasks.

\section*{Ethical Considerations}

One of the contributions is a multi-genre benchmark for semantic segmentation, so we pay particular attention to copyright, licensing, and the rights of original content creators. All material in the released benchmark is either (i) redistributed under terms that explicitly permit research use, or (ii) represented only via derived annotations and references, without including the protected source text.
For OPUS News Commentary,\footnote{\url{https://opus.nlpl.eu/News-Commentary}} we follow the corpus terms of use and the citation requirements of \citet{tiedemann-2012-parallel}, and we include the original MSA texts, which are already distributed for research purposes. Dialectal conversations from Rewayat are drawn from publicly available web sources; we only use modest-length excerpts and, to the best of our knowledge, found no terms restricting non-commercial research redistribution. The MGB-5 Moroccan broadcast transcripts were available online at the time of use without explicit additional licensing terms; we cite both the original challenge paper \citep{ali-mgb-5-2019} and the Hugging Face release.\footnote{\url{https://huggingface.co/datasets/ArabicSpeech/MGB-5}} For LDC telephone conversations \citep{ldc2006t15,ldc2006t16,ldc2007t01}, which are covered by a paid license and specific usage conditions, we do \emph{not} redistribute any original text; our benchmark only contains document identifiers and segment boundary metadata, so users must obtain the underlying transcripts directly from LDC under their own license. For the Mixat podcasts \citep{al-ali-aldarmaki-2024-mixat}, the data owners grant research-only usage rights; accordingly, we release the corresponding transcriptions and annotations solely for non-commercial research and explicitly mirror this restriction in our dataset license. We will comply with any future takedown or correction requests from rights holders.

The benchmark and models may reflect stereotypes or biases present in the underlying corpora (e.g., in dialectal speech, novels, or news), and we do not attempt to remove or normalize such content beyond basic preprocessing.

\section*{Acknowledgements}

We would like to warmly thank Prof.\ Ted Briscoe for his constant support, thoughtful conversations, and insights. His guidance helped us sharpen the main idea of the paper and improve the way its contribution is presented.



\bibliography{custom}

\clearpage

\appendix


\section{Synthetic annotation prompt template}
\label{sec:appendix-syn-annotation-prompt}
\small
\begin{verbatim}
Split the conversation ({data_source}) in {language_clue} into sequential segments,
where each segment contains lines that discuss the same topic.

The conversation is given as JSONL, one utterance per line, and each line has a unique
integer line_id. Your task is to output ONLY a JSON array of segment objects with
the following structure:

[
  {
    "split_id": "sequential integer starting from 1",
    "topic": "precise topic description in Modern Standard Arabic (max 10 words)",
    "line_ids": "comma-separated list of line_ids in order (e.g., \"1,2,3\")"
  }
]

SEGMENTATION RULES:
1) A new segment starts only when there is a clear topic shift
   (change of main subject, task, or goal).
2) Do NOT start a new segment for:
   - simple speaker changes,
   - backchannels or short clarifications,
   - minor digressions that stay within the same overall topic.
3) Segments must be contiguous in terms of line_ids: no overlaps, no gaps.
   Within each segment, line_ids must be consecutive (e.g., "1,2,3" is valid;
   "1,3,4" is invalid).

COVERAGE RULES (MUST hold for the entire document):
1) The first segment must start at the smallest line_id in the conversation.
2) Each next segment must start at the line_id immediately following the last
   line_id of the previous segment.
3) Every line_id in the input must appear exactly once in exactly one segment.

TOPIC FIELD:
- Write "topic" in clear, concise Modern Standard Arabic.
- Describe what is actually discussed in the segment, not the speaker or style.
- Do NOT invent content that is not supported by the lines.

IMPORTANT:
- Output MUST be valid JSON.
- Do NOT include any explanation, comments, or text outside the JSON array.

Conversation in {language_clue}:
---------------------------------------------
{conversation_str}
---------------------------------------------
\end{verbatim}

\clearpage
\onecolumn

\section{Annotation Guidelines}
\label{sec:appendix_guidelines}

\small

\subsection*{\small Goal}

Assess the segmentation quality line by line.  
Each segment is shown in a different color.

\subsection*{\small Given}

\begin{itemize}
    \item Table with \texttt{topic}, \texttt{line\_id}, \texttt{text}, \texttt{speaker} (original part) columns.  
    The original part is not for modification.
    \item The table contains sequential color-coded text segments that correspond to coherent topics.
    \item ``Validation'' fields are provided to assess the quality of segmentation.
\end{itemize}

\subsection*{\small Your job}

Check each colored segment: all lines within a segment should discuss one coherent idea/topic.

\begin{itemize}
    \item \textbf{Within-Segment Validation}: ``Is this line off-topic?''
    \begin{itemize}
        \item Keep this as \textbf{0} if the line is about the same topic as the previous line.
        \item Change this to \textbf{1} (i.e. off-topic) if the line doesn’t correspond to the whole segment (e.g., there is a sudden shift to a different topic).
    \end{itemize}

    \item \textbf{Cross-Segment Validation}: ``Is the main topic the same as the previous segment?''
    \begin{itemize}
        \item Keep this as \textbf{0} if the topic is different (default).
        \item Change this to \textbf{1} if the main topic in this line is actually the same as the previous segment (i.e., this should not have been segmented).  
        If you set this to 1, you are saying that the whole new segment should be merged with the previous segment.
    \end{itemize}
\end{itemize}

\subsection*{\small Examples}

\paragraph{1) {\small {Within-segment validation \#1.}}}
Text in line 432 doesn’t correspond to a topic.

\medskip
\noindent

\begin{center}
\begin{tabular}{l c c p{6cm} c c}
\toprule
topic & topic same as prev (0/1) & line\_id & text & speaker & off topic (0/1) \\
\midrule
\multirow{4}{*}{vendor renewal} &
0 & 430 & We’ll renew for 12 months with a 30-day exit clause. & A & 0 \\
&  & 431 & Legal will review the indemnity language. & B & 0 \\
&  & 432 & I’m running late; parking took forever. & B & 1 \\
&  & 433 & Add a data-processing addendum for SOC 2. & A & 0 \\
\bottomrule
\end{tabular}
\end{center}

\paragraph{2){\small  {Within-segment validation \#2.}}}
Text in line 322 doesn’t correspond to a topic.

\begin{center}
\begin{tabular}{l c c p{6cm} c c}
\toprule
topic & topic same as prev (0/1) & line\_id & text & speaker & off topic (0/1) \\
\midrule
\multirow{4}{*}{hiring plan} & 0 & 320 & We need a senior backend engineer in Q2. & A & 0 \\
& & 321 & I’ll draft the JD and open the requisition. & B & 0 \\
& & 322 & Also, the vendor’s contract expires Friday. & A & 1 \\
& & 323 & The target salary band is Level 6. & B & 0 \\
\bottomrule
\end{tabular}
\end{center}

\paragraph{3) {\small {Cross-segment \& within-segment validation \#3.}}}

\begin{enumerate}
    \item (Cross-segment validation) Topic in line 403 is the same as the previous topic and it could be merged.
    \item (Within-segment validation) Text in line 405 doesn’t correspond to the topic.
\end{enumerate}

\begin{center}
\begin{tabular}{ l c c p{6cm} c c}
\toprule
topic & topic same as prev (0/1) & line\_id & text & speaker & off topic (0/1) \\
\midrule
\multirow{3}{*}{project timeline} & 0 & 400 & Prototype by May 12 and beta by June 30. & A & 0 \\
& & 401 & QA window is two weeks after beta. & B & 0 \\
& & 402 & Design sign-off must happen before QA starts. & A & 0 \\
\midrule
\multirow{3}{*}{delivery schedule} &1 & 403 & I’ll circulate the calendar this afternoon. & B & 0 \\
& & 404 & We should add a one-week buffer for cert. & A & 0 \\
&  & 405 & Doorbell—give me ten seconds. & B & 1 \\
\bottomrule
\end{tabular}
\end{center}




\clearpage


\section{Silver vs.\ Golden Annotation Comparison}
\label{sec:appendix-silver-vs-gold-comparison}

\begin{table}[H]
\centering
\begin{tabular}{lr}
\toprule
\textbf{Source} & \textbf{Changes (\%)} \\
\midrule
OPUS (MSA)           & 31.02 \\
Rewayat (GL)       & 13.77 \\
MGB-5 (MR)          & 16.63 \\
LDC (GL)           & 17.14 \\
LDC (LV)      & 23.14 \\
LDC (IQ)          & 21.74 \\
Podcasts (CS) & 25.10 \\

\midrule
\textbf{Overall}     & \textbf{21.12} \\
\bottomrule
\end{tabular}
\caption{Percentage of annotation changes when comparing silver annotations against gold annotations across subsets. Higher values indicate that a larger proportion of silver annotations had to be revised.}
\label{tab:silver-vs-gold-comparison}
\end{table}

Table~\ref{tab:silver-vs-gold-comparison} shows that, overall, 21\% of silver annotations were changed during gold annotation. However, the amount of revision varies substantially across sources. The highest change rate is observed for \textit{OPUS (MSA)} is 31\%, followed by \textit{Podcasts (Code-Switching)} -- 25\% and \textit{LDC (Levantine)} -- 23\%. In contrast, the lowest change rates are found for \textit{Rewayat (Gulf)} is 14\%, \textit{MGB-5 (Moroccan)} , and \textit{LDC (Gulf)} -- 17\%).

In summary, these results show that about 1 in 5 parts of silver annotations were changed by human annotators in the final version. The rest remains unchanged.

\clearpage


\section{Segmentation prompt template}
\label{sec:appendix-segmentation-prompt}

\begin{verbatim}
Split the conversation ({{ data_source }}) in {{ language_clue }} into sequential segments,
where each segment contains lines that discuss the same topic.

The conversation is given as JSONL, one utterance per line, and each line has a unique
integer line_id. Your task is to output ONLY a JSON array of segment objects with
the following structure:

[
  {
    "split_id": "sequential integer starting from 1",
    "line_ids": "comma-separated list of line_ids in order (e.g., \"1,2,3\")"
  },
  ...
]

SEGMENTATION RULES:
1) A new segment starts only when there is a clear topic shift
   (change of main subject, task, or goal).
2) Do NOT start a new segment for:
   - simple speaker changes,
   - backchannels or short clarifications,
   - minor digressions that stay within the same overall topic.
3) Segments must be contiguous in terms of line_ids: no overlaps, no gaps.
   Within each segment, line_ids must be consecutive (e.g., "1,2,3" is valid;
   "1,3,4" is invalid).

COVERAGE RULES (MUST hold for the entire document):
1) The first segment must start at the smallest line_id in the conversation.
2) Each next segment must start at the line_id immediately following the last
   line_id of the previous segment.
3) Every line_id in the input must appear exactly once in exactly one segment.


IMPORTANT:
- Output MUST be valid JSON.
- Do NOT include any explanation, comments, or text outside the JSON array.

Conversation in {{ language_clue }}:
---------------------------------------------
{{ conversation_str }}
---------------------------------------------

\end{verbatim}

\clearpage

\onecolumn


\section{Segmentation restoration prompt template}
\label{sec:appendix-restoration-prompt}

\begin{verbatim}
Restore missing segment boundaries inside each DRAFT BLOCK for

{{ data_source }} ({{ language_clue }}).

You are given a conversation as JSONL utterances, grouped into DRAFT BLOCKS.
Each utterance has a unique integer "line_id".
DRAFT BLOCK outer borders are correct, but some internal segment borders may be missing.
Your job is to produce the FINAL segmentation by ADDING missing borders ONLY.

IMPORTANT RESTRICTION:
- You may ONLY split INSIDE a single block.
- You MUST NOT merge or move lines across blocks.
- Every block border must remain a segment border in the final output:
  • the first line_id of each block MUST be the first line_id of some output segment,
  • the last line_id of each block MUST be the last line_id of some output segment.

OUTPUT (ONLY):
Return ONLY a valid JSON array. Each element is a segment object with:
- "split_id": sequential integer starting from 1
- "line_ids": a comma-separated string of line_ids in increasing order

Example output format (note: JSON array, no extra text):
[
  {"split_id": 1, "line_ids": "2,3"},
  {"split_id": 2, "line_ids": "4,5,6,7"}
]

SEGMENTATION RULES:
1) Start a new segment ONLY when there is a clear topic shift
   (change of main subject / goal / task).
2) Do NOT start a new segment for:
   - speaker changes,
   - backchannels,
   - short clarifications,
   - minor digressions that stay within the same overall topic.

SEQUENTIAL LINE_ID RULE (MUST HOLD FOR EVERY SEGMENT):
- Within each segment, "line_ids" MUST be a sequence of consecutive integers (step = 1).
  Examples:
  - Valid:  "4,5,6,7"
  - Invalid: "4,6,7"  (missing 5)
  - Invalid: "7,6,5"  (not increasing)

COVERAGE RULES (MUST HOLD FOR THE ENTIRE RESPONSE):
1) ALL input line_ids across ALL blocks MUST be covered.
2) Each line_id MUST appear exactly once in exactly one output segment.
3) No gaps and no overlaps:
   - The first output segment MUST start with the smallest line_id in DRAFT BLOCK 1.
   - The last output segment MUST end with the largest line_id in DRAFT BLOCK N.
   - Across the entire output, segments must connect end-to-start:
     if one segment ends at line_id X, the next segment MUST start at line_id X+1.

FINAL CHECK BEFORE YOU OUTPUT:
- Do your segments cover every line_id shown in the blocks exactly once?
- Are line_ids consecutive within each segment?
- Did you preserve every DRAFT BLOCK border?

Blocks (each line is an utterance JSON object):
===== DRAFT BLOCK 1 =====
{{ draft_block_1_jsonl }}
===== DRAFT BLOCK 2 =====
{{ draft_block_2_jsonl }}
...
===== DRAFT BLOCK N =====
{{ draft_block_n_jsonl }}

\end{verbatim}

\clearpage

\section{Data Ablation Study}
\label{sec:appendix-data-ablation-study}
\begin{table}[H]
\centering
\footnotesize
\renewcommand{\arraystretch}{1.1}
\setlength{\tabcolsep}{2.5pt}
\scalebox{0.9}{
\begin{tabular}{lcccccccccccccccccc}
\toprule
\textbf{Method}
& \multicolumn{3}{c}{\textbf{OPUS}}
& \multicolumn{3}{c}{\textbf{Rewayat}}
& \multicolumn{3}{c}{\textbf{MGB-5}}
& \multicolumn{3}{c}{\textbf{LDC}}
& \multicolumn{3}{c}{\textbf{Podcasts}}
& \multicolumn{3}{c}{\textbf{Overall}} \\
\cmidrule(lr){2-4}\cmidrule(lr){5-7}\cmidrule(lr){8-10}\cmidrule(lr){11-13}\cmidrule(lr){14-16}\cmidrule(lr){17-19}
& F1($\uparrow$) & Pk($\downarrow$) & WD($\downarrow$)
& F1($\uparrow$) & Pk($\downarrow$) & WD($\downarrow$)
& F1($\uparrow$) & Pk($\downarrow$) & WD($\downarrow$)
& F1($\uparrow$) & Pk($\downarrow$) & WD($\downarrow$)
& F1($\uparrow$) & Pk($\downarrow$) & WD($\downarrow$)
& F1($\uparrow$) & Pk($\downarrow$) & WD($\downarrow$) \\
\midrule
All train data 
& \textbf{0.60} & \textbf{0.43} & \textbf{0.46}
& \textbf{0.52} & \textbf{0.43} & \textbf{0.43}
& \textbf{0.53} & \textbf{0.40} & \textbf{0.40}
& \textbf{0.50} & \textbf{0.41} & \textbf{0.39}
& \textbf{0.48} & \textbf{0.52} & \textbf{0.53}
& \textbf{0.52} & \textbf{0.43} & \textbf{0.43} \\
w/o OPUS 
& 0.52 & 0.52 & 0.56
& 0.51 & 0.44 & 0.45
& 0.50 & 0.43 & 0.47
& 0.49 & 0.42 & 0.41
& 0.41 & 0.53 & 0.63
& 0.49 & 0.47 & 0.50 \\
w/o Rewayat 
& \textbf{0.60} & \textbf{0.43} & \textbf{0.46}
& 0.45 & 0.50 & 0.52
& 0.52 & 0.41 & 0.42
& 0.49 & 0.42 & 0.40
& 0.47 & 0.53 & 0.54
& 0.51 & 0.46 & 0.47 \\
w/o MGB-5 
& \textbf{0.60} & \textbf{0.43} & \textbf{0.46}
& \textbf{0.52} & \textbf{0.43} & \textbf{0.43}
& 0.46 & 0.48 & 0.50
& \textbf{0.50} & 0.42 & 0.42
& \textbf{0.48} & \textbf{0.52} & \textbf{0.53}
& 0.51 & 0.46 & 0.47 \\
w/o LDC 
& \textbf{0.60} & 0.44 & 0.47
& 0.51 & 0.44 & 0.44
& 0.48 & 0.46 & 0.46
& 0.49 & 0.49 & 0.51
& 0.47 & 0.53 & 0.55
& 0.51 & 0.47 & 0.49 \\
w/o Podcasts 
& 0.59 & 0.44 & 0.47
& 0.51 & 0.44 & 0.44
& 0.52 & 0.41 & 0.41
& 0.49 & 0.42 & 0.40
& 0.46 & 0.58 & 0.60
& 0.51 & 0.46 & 0.46 \\
\bottomrule
\end{tabular}
}
\caption{Data ablation study. Each row removes one training subset. Performance measured on test dataset.}
\label{tab:data_ablation_study}
\end{table}

According to Table~\ref{tab:data_ablation_study}, a final model trained on all sources demonstrates the strongest and most stable overall performance, with the best scores across all three metrics ({\textbf {F1}}, {\textbf {Pk}}, {\textbf {WD}}). Removing any single source consistently lowers the overall result and mostly causes degradation in the corresponding evaluation subset. It indicates that each source contributes complementary information rather than redundant information. The final model benefits from the diversity of the full multi-genre training. In the table, bold marks the best value within each column: the highest value for {\textbf {F1}} and the lowest values for {\textbf {Pk}} and {\textbf {WD}}.

\end{document}